\documentclass[lettersize,journal]{IEEEtran}
\usepackage{amsmath,amsfonts}
\usepackage{algorithmic}
\usepackage{algorithm}
\usepackage{array}
\usepackage[caption=false,font=normalsize,labelfont=sf,textfont=sf]{subfig}
\usepackage{textcomp}
\usepackage{stfloats}
\usepackage{url}
\usepackage{verbatim}
\usepackage{graphicx}
\usepackage{cite}
\usepackage{amssymb}
\usepackage{bm}
\usepackage{longtable}
\usepackage{multirow}
\usepackage{arydshln} 
\usepackage{subfloat}
\usepackage{ulem}
\begin{document}

\title{Learning Cross-modality Information Bottleneck Representation for Heterogeneous Person Re-Identification}

\author{Haichao Shi,~\IEEEmembership{Member,~IEEE,}
		Mandi Luo,
		Xiao-Yu Zhang,~\IEEEmembership{Senior Member,~IEEE,}
		Ran He,~\IEEEmembership{Senior Member,~IEEE}
\thanks{H. Shi and X.-Y. Zhang are with the Institute of Information Engineering, Chinese Academy of Sciences, Beijing, China, 100093. (e-mail: shihaichao@iie.ac.cn; zhangxiaoyu@iie.ac.cn)}
\thanks{M. Luo is with Technology and Standards Research Institute, China Academy of Information and Communications Technology, Beijing, China, 100083. (e-mail: luomandi@caict.ac.cn)}
\thanks{R. He is with the Center for Research on Intelligent Perception and Computing, National Laboratory of Pattern Recognition, Institute of Automation, Chinese Academy of Sciences, CAS Center for Excellence in Brain Science and Intelligence Technology, Beijing, China, 100190. (e-mail: rhe@nlpr.ia.ac.cn)}
}

\markboth{Journal of \LaTeX\ Class Files,~Vol.~14, No.~8, August~2021}%
{Shell \MakeLowercase{\textit{et al.}}: A Sample Article Using IEEEtran.cls for IEEE Journals}


\maketitle

\begin{abstract}
Visible-Infrared person re-identification (VI-ReID) is an important and challenging task in intelligent video surveillance. Existing methods mainly focus on learning a shared feature space to reduce the modality discrepancy between visible and infrared modalities, which still leave two problems underexplored: \textbf{information redundancy} and \textbf{modality complementarity}. To this end, properly eliminating the identity-irrelevant information as well as making up for the modality-specific information are critical and remains a challenging endeavor. To tackle the above problems, we present a novel mutual information and modality consensus network, namely CMInfoNet, to extract modality-invariant identity features with the most representative information and reduce the redundancies. The key insight of our method is to find an optimal representation to capture more identity-relevant information and compress the irrelevant parts by optimizing a mutual information bottleneck trade-off. Besides, we propose an automatically search strategy to find the most prominent parts that identify the pedestrians. To eliminate the cross- and intra-modality variations, we also devise a modality consensus module to align the visible and infrared modalities for task-specific guidance. Moreover, the global-local feature representations can also be acquired for key parts discrimination. Experimental results on four benchmarks, i.e., SYSU-MM01, RegDB, Occluded-DukeMTMC, Occluded-REID, Partial-REID and Partial\_iLIDS dataset, have demonstrated the effectiveness of CMInfoNet.
\end{abstract}


\section{Introduction}
\IEEEPARstart{P}{erson} re-identification has been studied for several years, which is thought to be a specific person retrieval problem across non-overlapping cameras~\cite{1,agw,2,3}. Recent years, with the increasing frequency of malicious incidents at night-time, it has become an urgent problem to improve the pedestrian retrieval accuracy at night-time. The infrared cameras are thus applied to capture infrared images, which raises important visible-infrared person re-identification problems. Most of the existing methods focus on single-modality, which generally focuses on feature learning~\cite{featurelearning1,featurelearning2} and metric learning~\cite{metriclearning1,metriclearning2}. However, these methods cannot adapt well to illumination variations in real-world scenarios~\cite{Wu}. Therefore, identifying images captured by dual-mode cameras, i.e., visible and infrared cameras, has attracted much attention in the video surveillance community~\cite{cmgan,ssft}.

\begin{figure}[tbp]
\centering
\includegraphics[width=1\linewidth]{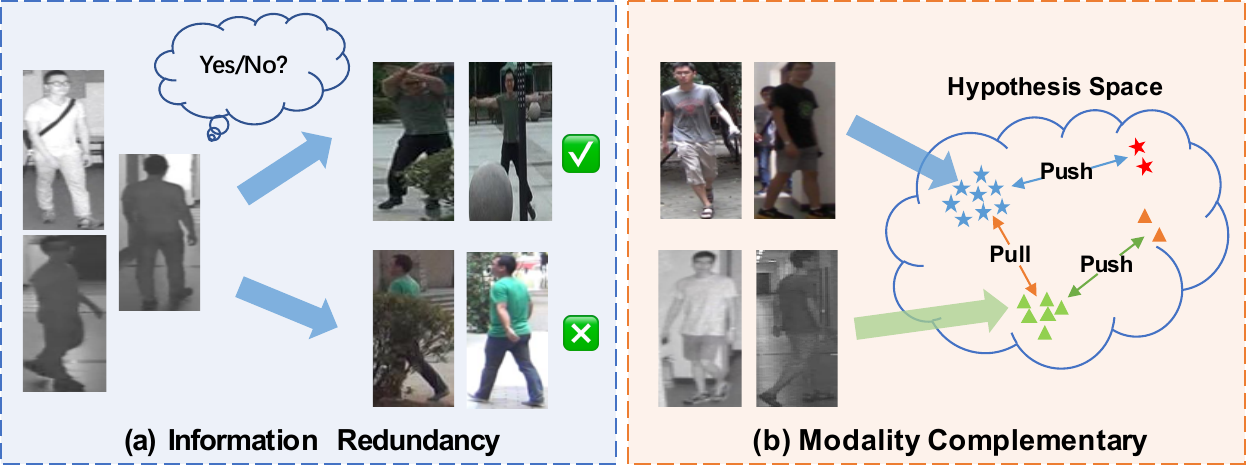}
\caption{Illustration of the two problems. \textbf{Information Redundancy}. The matching process will be affected by occlusion, illumination, etc., leading to inevitable misidentifications. \textbf{Modality Complementary}. Though mapped into the same feature space, modality-specific information is ignored without modal complement.}
\label{fig1}
\end{figure}

\begin{figure*}[tbp]
\centering
\includegraphics[width=1\linewidth]{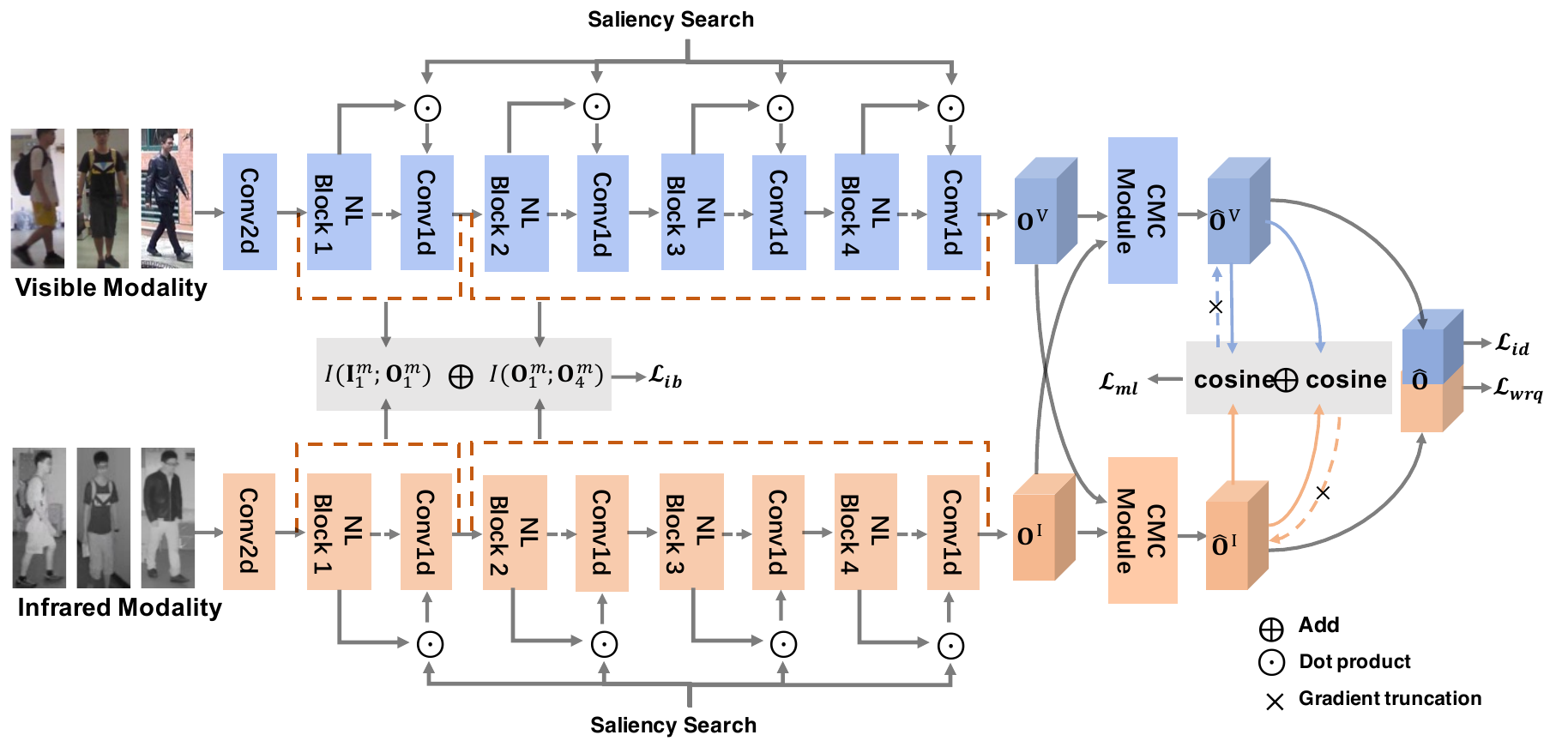}
\caption{Illustration of CMInfoNet, which takes two-stream modalities as inputs (better viewed in color). Note that `NL Block*' represents the $*$-th non-local block, guided by the mutual information bottleneck (MIB) for optimal identity information. The CMC module denotes the cross-modality consensus module for modality re-calibration.}
\label{fig2}
\end{figure*}

While dealing with VI-ReID, the challenges that existing methods are faced with are summarized as three-fold: \textbf{1) Information redundancy}, as shown in Fig.~\ref{fig1} (a). The visible and infrared images are with large modality gaps. Most of the existing VI-ReID methods ignore the intrinsically distinct characteristics, making it even harder to align images of the same identity. Prior methods adopt generative adversarial networks (GANs)~\cite{cmgan,aligngan} to eliminate the adverse effect of large modality gaps. However, the generation may have large color and texture variations compared with original probes, making it difficult to select the correct target that is generated for ReID. Besides, the matching process is often interferenced by external redundancies~\cite{sada,cocas} (e.g., occlusions, background, multi-views, etc.), which degrades the matching accuracy. \textbf{2) Modality complementarity}, as shown in Fig.~\ref{fig1} (b). To narrow the modality gap, most existing methods are devoted to learning a modality-shared feature embedding through mapping the features into a common feature space~\cite{Wu,Ye1,Ye2}. The specific information of different modalities is regarded as redundancy and is eliminated, ignoring the invariant features. However, the specific information plays an important role in conventional ReID methods. By means of mapping different modalities into the same feature space, the discrimination ability of modality-specific information will be neglected. Moreover, each modality is processed without attending to the other modality when mapped into the same feature space. \textbf{3) Small dataset}. The scale of VI-ReID datasets is rather small compared with that of other ReID datasets. For instance, the largest and newest dataset in VI-ReID, namely SYSU-MM01~\cite{Wu}, contains 395 identities with 34,167 images for training in total. There are approximately 126,000 images of 4,100 identities in the training set of the commonly used ReID dataset MSMT17~\cite{msmt}, which is notably larger than the SYSU-MM01 dataset. Training models with small datasets is more likely to suffer from the overfitting problem.

In this paper, we focus on improving the most expressive representation with cross-modality consensus learning. A novel mutual information-based cross-modality consensus network is proposed, namely CMInfoNet. We aim at learning the optimal representation from the identity and identity-irrelevant perspective. Inspired by the information theoretic principle, we optimize the mutual information bottleneck (MIB) trade-off to reduce the information redundancies and mitigate the overfitting and false correlation problems caused by small datasets. Inspired by the Neural Architecture Search (NAS) strategy, we propose an automatically search strategy to distinguish the saliency regions of pedestrians for improving the model efficiency and precision. Moreover, we devise a modality consensus module to align the visible and infrared modalities for task-specific guidance. The modality consensus module takes one main modality and one auxiliary modality as inputs, which are designed to attend to each other for modality complementary. Both the main and auxiliary modalities are integrated to explore the cross- and intra-modality variations. Meanwhile, a modality contrastive loss is designed to learn from each modality and explore the inter-modality consistency. Each single modality is regarded as calibration for the other. Inspired by Ye et al.~\cite{agw}, we also promote the WRT as weighted regularization quadruplet (WRQ) loss. 

The contributions of our work are summarized as follows:

\begin{itemize}
    \item We adopt the mutual information bottleneck to remove the identity-irrelevant redundancies and extract the minimal sufficient information of identities. Besides, the information bottleneck principle is beneficial to the overfitting and false correlation problems caused by small-scale datasets.
    \item We propose a saliency search algorithm to optimize the pedestrian representation to improve both the efficiency and the precision. 
    \item We introduce a cross-modality consensus module to explore the modality invariant identity features as well as align the features among different modalities for heterogeneous person re-identification.
    \item We devise a modality contrastive loss to maintain the inter-modality consistency, which preserves the modality-aware information when extracting features among different modalities.
    \item Extensive experiments have demonstrated the effectiveness and flexibility of our proposed method.
\end{itemize}

\section{Related Work}
In this section, we briefly describe the representative works related to ours, including single-modality person re-identification, visible-infrared person re-identification, neural architecture search and information bottleneck.

\subsection{Single-modality Person Re-Identification}
The goal of single-modality person re-identification is to match the pedestrian images obtained from non-overlapping visible cameras. To date in the literature, there are three main categories of approaches in single-modality person re-identification, i.e., the handcrafted descriptor based methods, the metric learning based methods and the deep learning based methods. The handcrafted descriptor based methods~\cite{metriclearning1,handcrafted1} generally adopt the Bag-of-Words (BoW) model to extract local features and match the global features. The metric learning based methods~\cite{relativedis,effpsd} generally learn a Mahalanobis distance function or projection matrix to optimize the feature representations. More recently, existing works~\cite{jdgl,Evolutionary,dsr,Beyond} have demonstrated superior performance on the widely-used benchmark datasets with the leverage of deep learning techniques. Among these methods, some methods~\cite{rank1,rank2} address the person re-identification problems as ranking problems to improve the retrieval performance. Some methods put main emphasis on learning the variants of widely-used loss functions~\cite{ppr,sona,abdnet} in person re-identification, i.e., identity loss, verification loss and triplet loss. Other methods~\cite{iida,csa} attempt to leverage the domain adaptation methods to regard the images from each visible camera as an independent domain. However, most cameras switch the visible mode to infrared mode during the night time, which impeds the performance of single modality re-identification due to the large cross-modality discrepancies.

\subsection{Visible-Infrared Person Re-Identification}
To reduce the cross-modality discrepancies, the visible-infrared person re-identification (VI-ReID) methods have been proposed to address the pedestrian retrieval problems captured from different cameras, i.e., visible and infrared cameras. The pioneer work~\cite{Wu} proposes a deep zero-padding framework, which aims to learn the modality sharable features. Besides, a new VI-ReID dataset namely SYSU-MM01 is also proposed. To further strengthen the pedestrian feature representation, researchers begin to employ the two-stream networks to model both the modality-shared and modality-specific features~\cite{agw,ddaa,bdcc}. Under this framework, the intra- and cross-modality variations can be addressed simultaneously. Besides, cm-SSFT~\cite{cmpr} integrates the transfer learning strategy to learn the modality-shared and modality-specific characteristics. In~\cite{sim}, a novel similarity inference metric is proposed to explore the intra-modality sample similarities to narrow the discrepancy of different modalities. MACE~\cite{mace} adopts the ensemble learning to model the modality-discrepancy in both feature level and classifier level. The XIV model~\cite{xmodality} is proposed to utilize an auxiliary X modality as an assistant and reformulate the VI-ReID as an X-Infrared-Visible three-mode learning problem. With the development of the generative models, the generative adversarial networks have been applied to generate data and mitigate the modality discrepancy. Specifically, AlignGAN~\cite{wang2019rgb} is proposed to jointly exploit the pixel and feature alignment with an end-to-end alignment generative adversarial network. To alleviate the problem of the lack of discriminative information and the difficulty to learn a robust metric for cross-modality retrieval, the cmGAN~\cite{dai2018cross} is proposed to re-identify the same person across different modalities. The pipeline of generating cross-modality paired-images~\cite{jsia} is also applied to perform the instance-level alignments and achieve finer-grained retrieval.

\subsection{Neural Architecture Search}
Recently, the deep learning techniques have been applied to various fields and achieved successful performance, such as computer vision, natural language processing, etc, which benefits from the emergence of the neural network structures, i.e., ResNet~\cite{resnet}, DenseNet~\cite{densenet}, etc. However, it is labourious and time-consuming to design a neural network with high superiority performance. The Neural Architecture Search (NAS)~\cite{dart} has been proposed to automatically design the neural networks, which are based on the sample sets through specific algorithms. Existing NAS methods are divided into two categories, i.e., micro search methods~\cite{pc-darts,srna,pdas} and macro search methods~\cite{FBNet,EfficientNet,ProxylessNAS,dnna}. The NAS have also been utilized in various fields, such as the image classification~\cite{DBLP:conf/aaai/RealAHL19}, semantic segmentation~\cite{DBLP:conf/iclr/ChenGLZLW20} and object detection~\cite{DBLP:conf/iclr/LiangLGSWYO20}. These methods mainly focus on the single-modality tasks, which do not take the cross-modality properties into account. The modality discrepancy need to be mitigated to improve the pedestrian retrieval performance. Also inspired by the characteristics of the NAS, we propose the cross-modality feature search algorithm to extract the saliency regions of pedestrians automatically.

\subsection{Information Bottleneck}
The information bottleneck (IB) is firstly introduced in~\cite{ib1} as an information theory, which provides a fundamental bound between the input compression and the target output information. The IB aims to extract the best trade-off between the model complexity and precision within a random variable and the observed relevant variable. In 2017, the Deep VIB~\cite{ib2} was proposed with the variational approximation to the information bottleneck, which led the IB to deep learning field for the first time. In the era of deep learning, the IBs has been widely used in many areas~\cite{ib3,ib4}, i.e., the graph neural networks~\cite{ib8}, the reinforcement learning~\cite{ib6,ib7}, and so on. The IBs has also achieved wide attention in the domain adaptation problems. In~\cite{ib9}, Peng et al. introduced the information bottleneck as a regularization to constrain the information flow in the discriminator, which has effectively improved the performance of Generative Adversarial Networks (GANs)~\cite{GAN} in the task of image generation. To alleviate the feature alignment in the task of unsupervised semantic segmentation, the significance-aware information bottleneck~\cite{ib10} is proposed to stabilize the adversarial training. In our work, we also leverage the IBs as a trade-off mechanism to address the domain discrepancy in VI-ReID to narrow the gap between the visible and infrared modalities.

\section{Methodology}
As shown in Fig.~\ref{fig2}, we illustrate the overview of CMInfoNet. The visible and infrared images are firstly fed into the two-stream framework to extract modality-invariant features, which is guided by the mutual information bottleneck (MIB). Meanwhile, the saliency search module is equipped with the non-local block to search the distinct characteristics of pedestrians automatically. Both the mutual information bottleneck and the saliency search module are responsible for the model efficiency and accuracy. Then, the cross-modality consensus (CMC) module aims to make the modality complement each other within a dual form of modalities, namely main modality and auxiliary modality. Then we present a modality contrastive learning strategy to re-calibrate the representations of each modality using information from different modality perspectives. The cross-modality identity representation is improved. The component details are discussed in the following parts.

Specifically, we denote the input as $\bm{\mathrm{I}}^m_n \in \mathbb{R}^{C_{in}\times W \times H}$ for each non-local block $n$. The corresponding output is denoted as $\bm{\mathrm{O}}^m_n \in \mathbb{R}^{C_{out}\times W' \times H'}$, where $m \in \{\mathrm{V},\mathrm{I}\}$, $\mathrm{V}$ and $\mathrm{I}$ represent the visible and infrared modalities, respectively. $C_{in/out}$, $W$, and $H$ are denoted as the input/output number of channels, width, and height, respectively. To make the whole feature extraction process involved, we select the output of the first and the last non-local block as the intermediate and final representations, denoted as $\bm{\mathrm{O}}_1^m$ and $\bm{\mathrm{O}}_4^m$, respectively.

\subsection{Saliency Search Module}
The saliency search module is responsible for optimizing the extracted features in the searched space, which is different from the topology architecture optimization of common NAS methods. We perform the pixel-level search among the output feature map of each non-local block. As shown in Fig.~\ref{fig2}, we perform different saliency search operation corresponding to different modalities. Specifically, the searched features are with the same dimension of the original output of each non-local block, which is $72 \times 36$, $36 \times 18$, $18 \times 9$, $9 \times 5$, respectively. Then the searched features are activated by a sigmoid function, which indicates the weight of a specific position in the extraction process. Therefore, the parameterized output activation map is calculated as follows.
\begin{equation}
\bm{\mathrm{O}}_1^m = \sigma(\bm{\mathrm{O}}_{N_1^{*}}^m) \odot \bm{\mathrm{O}}_{N_1}^m,
\end{equation}
where $*$ indicates the feature is drawn from the searched space, $N_1$ represents the output of the first non-local block, $\odot$ represents the Hadamard product.

Since the common NAS methods mainly focus on the optimization of network architecture in discrete space, which is not derivable. Our saliency search module deals with the feature optimization problem in a continuous space, where we can direct leverage the gradient descent algorithm to optimize the searched features.

\begin{figure}[tbp]
\centering
\includegraphics[width=1\linewidth]{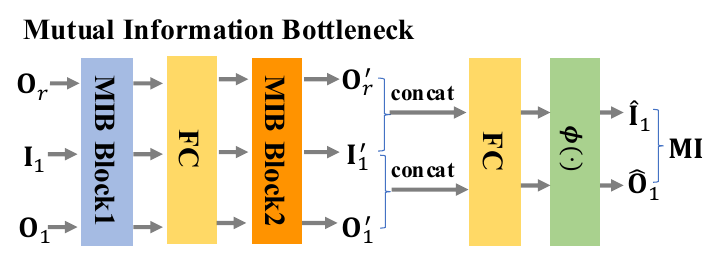}
\caption{Illustration of mutual information bottleneck module.}
\label{fig-ib}
\end{figure}

\subsection{Mutual Informative Modality Bottleneck}
Our goal is elaborated from two-fold: Firstly, we aim to learn the most expressive representations and reduce the information redundancies to adapt the VI-ReID, which can be further regarded as compressing the identity-irrelevant information. Secondly, it is noted that existing datasets used for VI-ReID are often in small-scale, leading to overfitting and false correlation problems. To tackle the problems, we introduce a mutual information bottleneck to guide the feature extraction process for more informative features. We adopt a generic framework namely AGW~\cite{agw} as the baseline feature extractor, which adopts the non-local based ResNet50 architecture. Fig.~\ref{fig-ib} illustrates an example of the detailed structure of MIB. Given input feature maps $\bm{\mathrm{I}}_1$, $\bm{\mathrm{O}}_1$ and the randomly shuffled $\bm{\mathrm{O}}_1$ as $\bm{\mathrm{O}}_r$, we can obtain the intermediate features through two MIB blocks. The concatenated features are then fed into an FC layer followed by a Softplus $\phi(\cdot)$. Finally the mutual information (MI) is calculated. The MIB block1 consists of three 3$\times$3 conv layers, of which the first two is followed by a BatchNormalization (BN) layer and a LeakyReLU activation function. The MIB block2 contains an FC, a BN and a LeakyReLU activation. On one hand, the pedestrian-related information can be specified to support the ReID. On the other hand, the pedestrian-unrelated parts could be the farthest suppressed. Thus, we can arrive at the MIB constraint: 

\begin{equation}
\mathcal{L}_{ib} = \min_{\bm{\mathrm{O}}_1^m} \alpha I(\bm{\mathrm{I}}_1^m; \bm{\mathrm{O}}_1^m) - \beta I(\bm{\mathrm{O}}_1^m; \bm{\mathrm{O}}_4^m),
\label{equ1}
\end{equation}
where $\alpha$ and $\beta$ are hyper-parameters that trade off the items. $I(\bm{\mathrm{I}}_1^m; \bm{\mathrm{O}}_1^m)$ is denoted as the mutual information that represents the relevance between $\bm{\mathrm{I}}_1^m$ and $\bm{\mathrm{O}}_1^m$. The goal is to minimize the mutual information of $\bm{\mathrm{O}}_1^m$ and $\bm{\mathrm{I}}_1^m$. Therefore, $\bm{\mathrm{O}}_1^m$ is as less relevant as possible to the $\bm{\mathrm{I}}_1^m$ and as more relevant as possible to the final pedestrian representation $\bm{\mathrm{O}}_4^m$ after optimizing Eqn. (\ref{equ1}). As the optimization objective $\bm{\mathrm{O}}_1^m$, the most discriminative identity-relevant information can be preserved.

The mutual information is defined by the Kullback-Leibler divergence~\cite{mine}:
\begin{equation}
I(\bm{\mathrm{I}}_1^m; \bm{\mathrm{O}}_1^m) = D_{KL}(\mathbb{P}_{\bm{\mathrm{I}}_1^m \bm{\mathrm{O}}_1^m} || \mathbb{P}_{\bm{\mathrm{I}}_1^m} \otimes \mathbb{P}_{\bm{\mathrm{O}}_1^m}),
\label{equ2}
\end{equation}
where $D_{KL}$ is defined as:
\begin{equation}
D_{KL}(\mathbb{P}||\mathbb{Q}):=\mathbb{E}_{\mathbb{P}}[\log\frac{d\mathbb{P}}{d\mathbb{Q}}].
\end{equation}
$\mathbb{P}_{\bm{\mathrm{I}}_1^m \bm{\mathrm{O}}_1^m}$ and $\mathbb{P}_{\bm{\mathrm{I}}_1^m} \otimes \mathbb{P}_{\bm{\mathrm{O}}_1^m}$ are denoted as the joint distributions and the dot product of the marginals, respectively. $\mathbb{P}$ is also absolutely continuous with respect to $\mathbb{Q}^2$. The dependence between $\bm{\mathrm{I}}_1^m$ and $\bm{\mathrm{O}}_1^m$ can be further optimized with larger KL-divergence between $\mathbb{P}_{\bm{\mathrm{I}}_1^m \bm{\mathrm{O}}_1^m}$ and $\mathbb{P}_{\bm{\mathrm{I}}_1^m} \otimes \mathbb{P}_{\bm{\mathrm{O}}_1^m}$.

However, as Eqn. (\ref{equ2}) is not optimized using neural networks, we further introduce a deep neural network with parameters $\theta \in \Theta$~\cite{mine} to exploit the distribution bound, which can be represented as $I(\bm{\mathrm{I}}_1^m; \bm{\mathrm{O}}_1^m)  \geq I_{\Theta}(\bm{\mathrm{I}}_1^m, \bm{\mathrm{O}}_1^m)$.

We adopt the Jensen-Shannon representation (JS) to be the neural information measure, which is stable in the neural network optimizing process. The estimated JS mutual information is defined as:
\begin{equation}
\begin{split}
I_{\Theta}(\bm{\mathrm{I}}_1^m, \bm{\mathrm{O}}^m_1) &= \sup_{\theta \in \Theta}\mathbb{E}_{\mathbb{P}_{\bm{\mathrm{I}}_1^m \bm{\mathrm{O}}_1^m}}[-\phi(-T_{\theta})] \\
&- \mathbb{E}_{\mathbb{P}_{\bm{\mathrm{I}}_1^m} \otimes \mathbb{P}_{\bm{\mathrm{O}}_1^m}}[\phi(T_{\theta})],
\end{split}
\label{equ4}
\end{equation}
where $\{T_{\theta}\}_{\theta \in \Theta}$ is denoted as the function set that is parameterized by a neural network. Thus, the mutual information can be maximized. $\phi(\cdot)$ represents a Softplus operation. In Eqn. (\ref{equ4}), the two expectations are optimized across all functions $T_{\theta}$, which can be finite after the operation.

Similarly, the second item in Eqn. (\ref{equ1}) is calculated by:
\begin{equation}
\begin{split}
I_{\Theta}(\bm{\mathrm{O}}_1^m, \bm{\mathrm{O}}_4^m) &= \sup_{\theta \in \Theta} \mathbb{E}_{\mathbb{P}_{\bm{\mathrm{O}}_1^m \bm{\mathrm{O}}_4^m}}[-\phi(-T_{\theta})] \\
&- \mathbb{E}_{\mathbb{P}_{\bm{\mathrm{O}}_1^m} \otimes \mathbb{P}_{\bm{\mathrm{O}}_4^m}}[\phi(T_{\theta})].
\end{split}
\label{equ5}
\end{equation}
Therefore, we can reformulate the Eqn. (\ref{equ1}) as:
\begin{equation}
\begin{split}
\mathcal{L}_{ib} = \min_{\bm{\mathrm{O}}_1^m} & \sup_{\theta \in \Theta} \alpha \mathbb{E}_{\mathbb{P}_{\bm{\mathrm{I}}_1^m \bm{\mathrm{O}}_1^m}}[-\phi(-T_{\theta})] -  \alpha \mathbb{E}_{\mathbb{P}_{\bm{\mathrm{I}}_1^m} \otimes \mathbb{P}_{\bm{\mathrm{O}}_1^m}}[\phi(T_{\theta})] \\
& -\beta  \mathbb{E}_{\mathbb{P}_{\bm{\mathrm{O}}_1^m \bm{\mathrm{O}}_4^m}}[-\phi(-T_{\theta})] + \beta \mathbb{E}_{\mathbb{P}_{\bm{\mathrm{O}}_1^m} \otimes \mathbb{P}_{\bm{\mathrm{O}}_4^m}}[\phi(T_{\theta})],
\end{split}
\label{equ6}
\end{equation}
where $\alpha$ and $\beta$ are set to 1 in our case. The parameters of mutual information bottleneck as well as the non-local based ResNet50 are optimized together.

\subsection{Cross-modality Consensus Module}
\begin{figure}[tbp]
\centering
\includegraphics[width=0.9\linewidth]{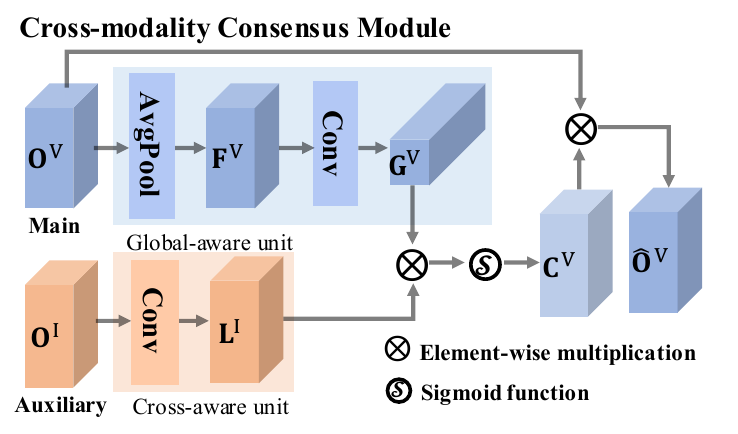}
\caption{Illustration of cross-modality consensus module.}
\label{fig-cmc}
\end{figure}

In this section, we elaborate the design of cross-modality consensus (CMC) module. The CMC module aims to filter out the task-irrelevant information of each modality and align the visible and infrared modalities for task-specific guidance. The CMC module takes two modalities as inputs, one is called main modality, which is responsible for exploring the global-modality information of current branch by a global-aware unit. The other is the auxiliary modality, aiming to distinguish the information redundancy and re-calibrate the representations by a cross-aware unit. Both two modalities are optimized to explore the cross- and intra-modality variations and complement the identity information. Fig.~\ref{fig-cmc} illustrates the detailed design of CMC module. The visible modality is regarded as the main modality and the infrared modality is considered as the auxiliary. The same workflow is performed when the roles of two modalities are exchanged.

\textbf{Global-aware unit.} Although most of the existing methods adopt ResNet as the backbone for pedestrian feature extraction, it is pretrained neither on large-scale pedestrian datasets nor on specific ReID datasets. These features could always contain some task-irrelevant redundancy that restricts the identification accuracy. Instead of directly fusing the main and auxiliary modality, we intend to filter out the task-irrelevant redundancy of the main modality and align the two modalities for invariant feature extraction. Inspired by the squeeze-and-excitation block~\cite{senet} and self-attention mechanism~\cite{self-attention}, we devise a channel-wise fusion mechanism to distinguish and filter out the task-irrelevant information of the main modality.

To be specific, given the extracted visible and infrared features, $\bm{\mathrm{O}}^{\mathrm{V}}$ and $\bm{\mathrm{O}}^{\mathrm{I}} \in \mathbb{R}^{C_{out} \times W' \times H'}$. We first squeeze the global information $\bm{\mathrm{O}}^{\mathrm{V}}$ into a channel-level feature $\bm{\mathrm{F}}^{\mathrm{V}} \in \mathbb{R}^{C_{out} \times 1 \times 1}$ via an average pooling (AVE) layer. Then a convolutional layer is adopted to further capture the channel-wise dependencies and produce the global modality-aware descriptor $\bm{\mathrm{G}}^{\mathrm{V}}$, which can be formulated as follows:
\begin{equation}
\bm{\mathrm{F}}^{\mathrm{V}} = \mathrm{AVE}(\bm{\mathrm{O}}^{\mathrm{V}}), \bm{\mathrm{G}}^{\mathrm{V}} = \mathrm{Conv}(\bm{\mathrm{F}}^{\mathrm{V}}).
\end{equation}

\textbf{Cross-aware unit.} As we know, the information from different modalities can be described from different perspectives, which can be leveraged to help identify the unique characteristic of each modality. Therefore, we leverage the infrared features $\bm{\mathrm{O}}^{\mathrm{I}}$ to be the auxiliary modality to capture cross modality-aware information and detect the task-irrelevant information of main modality. On the opposite of global-aware unit, the cross-modality local information is captured in cross-aware unit. A convolutional layer is also utilized to extract the embeddings of auxiliary modality to produce a local modality-aware representation $\bm{\mathrm{L}}^{\mathrm{I}}$.

Followed by the self-attention mechanism, we treat the global modality-aware descriptor $\bm{\mathrm{G}}^{\mathrm{V}}$ and local modality-aware descriptor $\bm{\mathrm{L}}^{\mathrm{I}}$ as `Query' and `Key', respectively. Then we multiply them to obtain the final channel-wise descriptor $\bm{\mathrm{C}}^{\mathrm{V}}$ for modality re-calibration. Finally, the task-irrelevant information redundancy can be filtered out.
\begin{equation}
\bm{\mathrm{C}}^{\mathrm{V}} = \bm{\mathrm{G}}^{\mathrm{V}} \otimes \bm{\mathrm{L}}^{\mathrm{I}}, \hat{\bm{\mathrm{O}}}^{\mathrm{V}} = \sigma(\bm{\mathrm{C}}^{\mathrm{V}})\otimes \bm{\mathrm{O}}^{\mathrm{V}},
\end{equation}
where $\otimes$ represents the element-wise multiplication operation and $\sigma(\cdot)$ is a sigmoid function. Similarly, when the infrared features are regarded as the main modality, the visible features are seen as auxiliary modality, we can obtain the calibrated identity features $\hat{\bm{\mathrm{O}}}^{\mathrm{I}}$ for infrared modality.

\subsection{Training Objectives}
In this section, we formulate the overall training objectives to better guide the cross-modality ReID process.

\textbf{Identity loss.} The identity loss $\mathcal{L}_{id}$~\cite{idloss} is conventional designed to learn the most discriminative representations for ReID.

\textbf{Weighted Regularization Quadruplet (WRQ) loss.} The Weighted Regularization Triplet (WRT) loss~\cite{agw} is proposed to optimize the relative distance between positive and negative pairs with an entire parameter-free strategy based on triplet loss. However, the triplet loss pays more emphasis on the relative distances between positive and negative pairs w.r.t the same probe images. This strategy suffers from a weaker generalization capability from training set to testing set, resulting in inferior performance. Therefore, we improve the WRT loss as WRQ loss, which pushes away negative pairs from positive pairs w.r.t different probe images. Compared with the WRT loss, the WRQ loss optimizes the output with a larger inter-class variation and a smaller intra-class variation. The WRQ loss $\mathcal{L}_{wrq}$ is represented as:

\begin{equation}
\begin{split}
\mathcal{L}_{wrq} = & {\sum}_{i,j,k} \log(1+\exp(\omega_i^p d_{ij}^p - \omega_i^n d_{ik}^n)) \\
 + & {\sum}_{i,j,k,l} \log(1+\exp(\omega_i^p d_{ij}^p - \omega_l^n d_{lk}^n)).
\end{split}
\label{equ9}
\end{equation}
\begin{equation}
\begin{split}
& \omega_i^p = \frac{\exp(d_{ij}^p)}{{\sum}_{d^p \in \mathcal{P}} \exp(d^p)}, \omega_i^n  = \frac{\exp(-d_{ik}^n)}{{\sum}_{d^n \in \mathcal{N}} \exp(-d^n)},\\
& \omega_l^n  = \frac{\exp(d_{lk}^n)}{{\sum}_{d^n \in \mathcal{N}} \exp(d^n)}, s_i = s_j, s_l \neq s_k, s_i \neq s_l, s_i \neq s_k,
\end{split}
\end{equation}
where $(i,j,k,l)$ denotes the quadruplet within each training batch. The quadruplet $(i,j,k,l)$ represents anchor, positive sample, negative sample and negative sample, respectively. $d_{ij}^p$ measures eucilidean distance between identity $i$ and $j$. $\mathcal{P}$ and $\mathcal{N}$ are the corresponding positive and negative set, respectively. $s_i$ refers to the identity ID. The second term in Eqn. (\ref{equ9}) considers the positive and negative images orders with different probes, which can further enlarge the inter-class variations.

\textbf{Modality contrastive loss.} Through the cross-modality consensus module, we can obtain the task-specific modality features. We further adopt a modality contrast pattern to make the two modalities learn from each other. The modality contrastive loss $\mathcal{L}_{ml}$ can be represented as follows:\begin{equation}
\mathcal{L}_{ml} = \lambda \Phi(\hat{\bm{\mathrm{O}}}^{\mathrm{V}}, \Delta(\hat{\bm{\mathrm{O}}}^{\mathrm{I}})) + (1-\lambda) \Phi(\hat{\bm{\mathrm{O}}}^{\mathrm{I}}, \Delta(\hat{\bm{\mathrm{O}}}^{\mathrm{V}})),
\end{equation}
where $\Phi(\cdot)$ is a similarity metric function and $\lambda$ is a hyperparameter. $\Delta$ represents a gradient truncation function.

\textbf{Information bottleneck loss.} The information bottleneck loss $\mathcal{L}_{ib}$ is used to extract more informative features, as depicted in Eqn. (\ref{equ6}). Therefore, we arrive at the final objective function for whole framework optimization:
\begin{equation}
\mathcal{L} = \mathcal{L}_{id} + \mathcal{L}_{wrq} + \lambda_{ml}\mathcal{L}_{ml} + \lambda_{ib}\mathcal{L}_{ib}.
\end{equation} 
where $\lambda_{ml}$ and $\lambda_{ib}$ are two trade-off parameters.

\begin{table}[tbp]
\small
\centering
\caption{Analysis of the effectiveness of MIB and CMC module on SYSU-MM01 dataset under \textit{all-search} mode. Rank-1 accuracy (\%), mAP (\%) and mINP (\%) are reported. Note that `Baseline' denotes pipeline of AGW, `MIB' is short for mutual information bottleneck and `CMC' represents cross-modality consensus module.}
\begin{tabular}{l|ccc}
\hline
Methods&Rank-1&mAP&mINP\\
\hline
Baseline & 48.83 & 47.96 & 36.07 \\
Baseline + MIB& 53.07 & 51.83 & 39.89 \\
Baseline + CMC& 55.16 & 54.39 & 41.72 \\
Baseline + MIB + CMC& 60.25 & 58.90 & 45.66\\
\hline
CMInfoNet*&\underline{63.46}&\underline{60.53}&\underline{47.27}\\
CMInfoNet$\star$&\textbf{64.89}&\textbf{61.36}&\textbf{48.13}\\
\hline
\end{tabular}
\label{table1}
\end{table}

\begin{table}[tbp]
\small
\centering
\caption{Comparisons of different kinds of combinations for the main and the auxiliary modality in CMC module on SYSU-MM01 dataset under the \textit{all-search} mode. Note that `Global' denotes the global average pooling operation is adopted after the convolution layer, while `Single' denotes only convolution operation.}
\begin{tabular}{p{0.6cm}<{\centering}c|p{0.6cm}<{\centering}p{0.6cm}<{\centering}p{0.6cm}<{\centering}|p{0.6cm}<{\centering}p{0.6cm}<{\centering}p{0.6cm}<{\centering}}
\hline
\multirow{2}*{Main}&\multirow{2}*{Auxiliary}&\multicolumn{3}{c}{Single-shot}&\multicolumn{3}{c}{Multi-shot}\\
\cline{3-8}
& &R1&mAP&mINP&R1&mAP&mINP\\
\hline
Single&Single& 64.95 & 62.36 & 48.39 & 71.64 & 55.53 & 18.62\\
Single&Global& 65.28 & 62.60 & 49.22 & 72.33 & 56.37 & 19.18\\
Global&Global& 65.06 & 62.53 & 48.75 & 71.97 & 56.02 & 18.95\\
Global&Single&\textbf{67.92}&\textbf{63.42}&\textbf{49.93}&\textbf{72.86}&\textbf{56.89}&\textbf{19.70}\\
\hline
\end{tabular}
\label{table2}
\end{table}

\section{Experiments}
In this section, we evaluate our method on two holistic publicly available datasets, i.e., SYSU-MM01~\cite{Wu} and RegDB~\cite{regdb}, and four partial re-id datasets, i.e., Occluded-DukeMTMC~\cite{Occluded-Duke}, Occluded-REID~\cite{Occluded-REID}, Partial-REID~\cite{Partial-REID} and PartialiLIDS~\cite{PartialiLIDS}.

\begin{table}[tbp]
\centering
\small
\caption{Comparison of different combinations of training objectives on SYSU-MM01 dataset under \textit{all-search} and \textit{single-shot} mode.}
\begin{tabular}{cccc|ccc}
\hline
$\mathcal{L}_{id}$&$\mathcal{L}_{wrq}$&$\mathcal{L}_{ml}$&$\mathcal{L}_{ib}$&Rank-1&mAP&mINP\\
\hline
\checkmark&-&-&-& 55.83 & 52.56 & 42.40 \\
\checkmark&\checkmark&-&-& 60.91 &59.62 & 43.28 \\
\checkmark&\checkmark&\checkmark&-& 63.78 & 60.78 & 46.52\\
\checkmark&\checkmark&-&\checkmark& 64.89 & 61.36 & 48.13 \\
\checkmark&\checkmark&\checkmark&\checkmark&\textbf{67.92}&\textbf{63.42}&\textbf{49.93}\\
\hline
\end{tabular}
\label{table3}
\end{table}

\begin{table}[tbp]
\centering
\small
\caption{Comparison of different similarity metrics of $\mathcal{L}_{ml}$ on SYSU-MM01 (\textit{all-search}) and RegDB (\textit{Visible to Infrared}) dataset.}
\begin{tabular}{c|ccc|ccc}
\hline
\multirow{2}*{Metric}&\multicolumn{3}{c|}{SYSU-MM01}&\multicolumn{3}{c}{RegDB}\\
\cline{2-7}
&Rank-1&mAP&mINP&Rank-1&mAP&mINP\\
\hline
JS&65.35 & 62.90 & 49.64& 85.89 & 82.25 & 68.22\\
MSE&66.71 & 63.18 & 49.80& 85.96 & 82.40 & 68.83\\
Cosine&\textbf{67.92}&\textbf{63.42}&\textbf{49.93} & \textbf{86.13} & \textbf{82.68} & \textbf{69.02} \\
\hline
\end{tabular}
\label{table4}
\end{table}

\subsection{Datasets and Evaluation Metrics}
\textbf{SYSU-MM01} dataset~\cite{Wu} is the largest benchmark dataset for VI-ReID, which is collected by 6 cameras, consisting of 4 visible and 2 infrared cameras. Specifically, it contains 22,258 visible images and 11,909 infrared images with 395 identities for training. The test set contains 96 persons, including 301 randomly selected visible images as gallery set and 3,803 infrared images for query. Meanwhile, it contains two different testing modes, all-search and indoor search modes. 

\textbf{RegDB} dataset~\cite{regdb} is collected by dual camera systems, which contains 412 identities with 206 identities for training and 206 identities for testing. Each identity has 10 images from a visible camera and 10 images from a thermal camera. There are also two evaluation modes, i.e., Visible to Thermal to search IR images from a Visible image and Thermal to Visible to search visible images from an infrared image. 

\textbf{Occluded-DukeMTMC} dataset~\cite{Occluded-Duke} consists of 15,618 training
images of 702 persons, 2,210 query images of 519 persons,
and 17,661 gallery images of 1,110 persons. It is the most
challenging occluded person ReID datasets due to the diverse scenes and distractions.

\textbf{Occluded-REID} dataset~\cite{Occluded-REID} is an occluded person ReID dataset captured by mobile cameras. It consists of 2,000 images belonging to 200 identities. Each identity has five full-body person images and five occluded person images with different viewpoints and different types of severe occlusions.

\textbf{Partial-REID} dataset~\cite{Partial-REID} is a specially designed ReID dataset that consists of occluded, partial, and holistic pedestrian images. It involves 600 images of 60 persons. We take the occluded query set and holistic galley set for experiments.

\begin{table*}[ht]
\small
\centering
\caption{Comparison on the SYSU-MM01 dataset with Rank-1, 10, 20 (\%) accuracy and mAP (\%) evaluation metrics.}
\begin{tabular}{p{2.6cm}<{\centering}|p{0.55cm}<{\centering}p{0.55cm}<{\centering}p{0.55cm}<{\centering}p{0.55cm}<{\centering}|p{0.55cm}<{\centering}p{0.55cm}<{\centering}p{0.55cm}<{\centering}p{0.55cm}<{\centering}|p{0.55cm}<{\centering}p{0.55cm}<{\centering}p{0.55cm}<{\centering}p{0.55cm}<{\centering}|p{0.55cm}<{\centering}p{0.55cm}<{\centering}p{0.55cm}<{\centering}p{0.55cm}<{\centering}}
\hline
\multirow{3}*{Method}&\multicolumn{8}{c|}{All Search}&\multicolumn{8}{c}{Indoor Search}\\
\cline{2-17}
&\multicolumn{4}{c}{Single-shot}&\multicolumn{4}{c|}{Multi-shot}&\multicolumn{4}{c}{Single-shot}&\multicolumn{4}{c}{Multi-shot}\\
\cline{2-17}
&R1&R10&R20&mAP&R1&R10&R20&mAP&R1&R10&R20&mAP&R1&R10&R20&mAP\\
\hline
Zero-Pad~\cite{Wu}&14.80&54.12&71.33&15.95&19.13&61.40&78.41&10.89&20.58&68.38&85.79&26.92&24.43&75.86&91.32&18.64\\
cmGAN~\cite{cmgan}&26.97&67.51&80.56&27.80&31.49&72.74&85.01&22.27&31.63&77.23&89.18&42.19&37.00&80.94&92.11&32.76\\
AlignGAN~\cite{aligngan}&42.4&85.0&93.7&40.7&51.5&89.4&95.7&33.9&45.9&87.6&94.4&54.3&57.1&92.7&97.4&45.3\\ 
JSIA-ReID~\cite{jsia}&38.1&80.7&89.9&36.9&45.1&85.7&93.8&29.5&43.8&86.2&94.2&52.9&52.7&91.1&96.4&42.7\\ 
MIR~\cite{msr}&37.35&83.40&93.34&38.11&43.86&86.94&95.68&30.48&39.64&89.29&97.66&50.88&46.56&93.57&98.80&40.08\\ 
MSTN~\cite{mace}&51.64&87.25&94.44&50.11&-&-&-&-&57.35&93.02&97.47&64.79&-&-&-&-\\ 
DDAG~\cite{ddag}&54.75&90.39&95.81&53.02&-&-&-&-&61.02&94.06&98.41&67.98&-&-&-&-\\ 
AGW~\cite{agw}&47.50&84.39&92.14&47.65&-&-&-&-&54.17&91.14&95.98&62.97&-&-&-&-\\ 
cm-SSFT~\cite{ssft}*&47.7&-&-&54.1&57.4&-&-&\textbf{59.1}&-&-&-&-&-&-&-&-\\ 
HAT~\cite{hat}&55.29&92.14&97.36&53.89&-&-&-&-&62.10&95.75&99.20&69.37&-&-&-&-\\ 
CMAlign~\cite{CMAlign}&55.41&-&-&54.14&-&-&-&-&58.46&-&-&66.33\\ 
NFS~\cite{nfs}&56.91&91.34&96.52&55.45&63.51&94.42&97.81&48.56&62.79&96.53&99.07&69.79&70.03&97.70&99.51&61.45\\ 
SMCL~\cite{SMCL}&\underline{67.39}&92.87&96.76&61.78&\underline{72.15}&90.66&94.32&54.93&68.84&96.55&98.77&75.56&\underline{79.57}&95.33&98.00&\underline{66.57}\\ 
CM-NAS~\cite{CM-NAS}&61.99&92.87&\underline{97.25}&60.02&68.68&\underline{94.92}&\underline{98.36}&53.45&67.01&\underline{97.02}&\underline{99.32}&72.95&76.48&\underline{98.68}&\textbf{99.91}&65.11\\ 
MCLNet~\cite{MCLNet}&65.40&\underline{93.33}&97.14&\underline{61.98}&-&-&-&-&\underline{72.56}&96.98&99.20&\underline{76.58}&-&-&-&-\\ 
\hline
CMInfoNet&\textbf{67.92}&\textbf{95.13}&\textbf{98.35}&\textbf{63.42}&\textbf{72.86}&\textbf{97.04}&\textbf{99.12}&\underline{56.89}&\textbf{73.22}&\textbf{97.06}&\textbf{99.50}&\textbf{76.81}&\textbf{79.98}&\textbf{98.72}&\underline{99.64}&\textbf{68.76}\\
\hline
\end{tabular}
\label{table5}
\end{table*}

\begin{table*}[ht]
\small
\centering
\caption{Comparison on the RegDB dataset with Rank-1, 10, 20 (\%) accuracy and mAP (\%) evaluation metrics.}
\begin{tabular}{c|cccc|cccc}
\hline
\multirow{2}*{Method}&\multicolumn{4}{c|}{Visible to Infrared}&\multicolumn{4}{c}{Infrared to Visible}\\
\cline{2-9}
&R1&R10&R20&mAP&R1&R10&R20&mAP\\
\hline
Zero-Pad~\cite{Wu}&17.75&34.21&44.35&18.90&16.63&34.68&44.25&17.82\\
AlignGAN~\cite{aligngan}&57.9&-&-&53.6&56.3&-&-&53.4\\ 
JSIA-ReID~\cite{jsia}&48.5&-&-&49.3&48.1&-&-&48.9\\ 
MSR~\cite{msr}&48.43&70.32&79.95&48.67&-&-&-&-\\ 
MSTN~\cite{mace}&72.37&88.40&93.59&69.09&72.12&88.07&93.07&68.57\\ 
DDAG~\cite{ddag}&69.34&86.19&91.49&63.46&68.06&85.15&90.31&61.80\\ 
AGW~\cite{agw}&70.05&86.21&91.55&66.37&70.49&87.12&91.84&65.90\\ 
cm-SSFT~\cite{ssft}&65.4&-&-&65.6&63.8&-&-&64.2\\ 
HAT~\cite{hat}&71.83&87.16&92.16&67.56&70.02&86.45&91.61&66.30\\ 
CMAlign~\cite{CMAlign}&74.17&-&-&67.64&72.43&-&-&65.46\\ 
NFS~\cite{nfs}&80.54&91.96&95.07&72.10&77.95&90.45&93.62&69.79\\ 
SMCL~\cite{SMCL}&83.93&-&-&79.83&\underline{83.05}&-&-&\underline{78.57}\\
CM-NAS~\cite{CM-NAS}&\underline{84.54}&\underline{95.18}&\underline{97.85}&\underline{80.32}&82.57&\underline{94.51}&\underline{97.37}&78.31\\ 
MCLNet~\cite{MCLNet}&80.31&92.70&96.03&73.07&75.93&90.93&94.59&69.49\\ 
\hline
CMInfoNet&\textbf{86.13}&\textbf{95.42}&\textbf{97.93}&\textbf{82.68}&\textbf{84.30}&\textbf{94.89}&\textbf{97.46}&\textbf{79.83}\\
\hline
\end{tabular}
\label{table6}
\end{table*}

\textbf{Partial\_iLIDS} dataset~\cite{PartialiLIDS} is a simulated partial re-id dataset based on the iLIDS dataset~\cite{PartialiLIDS}. The iLIDS dataset contains a total of 476 images of 119 people captured by multiple non-overlapping cameras. We conduct experiments on Partial-REID and Partial-iLIDS to demonstrate the effectiveness of our proposed method when facing the Partial Re-ID problem.

\textbf{Market-1501} dataset~\cite{handcrafted1} contains 12,936 training images of 751 persons, 19,732 query images and 3,368 gallery images of 750 persons captured from 6 cameras. It contains few of occluded person images. Since the Partial-REID, Partial\_iLIDS and Occluded-REID dataset do not have corresponding training set, the Market-1501 is only used for training the model to evaluate the above three occluded datasets as previous work do~\cite{dsr,sfr}.

\textbf{Evaluation Metrics.} The standard Cumulative Matching Characteristic (CMC), mean average precision (mAP) and mean inverse negative penalty~\cite{agw} (mINP) are used as evaluation metrics. Following~\cite{Wu}, we evaluate the results of SYSU-MM01 with offical code based on the average of 10 times repeated random split of gallery and probe set. Following~\cite{Ye1}, the results of RegDB are also based the average of 10 times repeated random split of training and testing sets.

\subsection{Implementation Details}
Firstly, we train the saliency search module to optimize $\bm{\mathrm{O}}_{N_1^{*}}^m$. The Adam optimizer is utilized with a learning rate of 1e-3. The searched model is then applied to the Re-ID model once the pixel-level optimization is finished. The CMInfoNet adopts a two-stream non-local based ResNet50 as the baseline feature extractor~\cite{agw}. The ResNet is pretrained on ImageNet. We set the batchsize to 32, which contains 16 visible and 16 infrared images from 4 identities. The input images are resized to 288 $\times$ 144. We use the Stochastic Gradient Descent (SGD) optimizer for optimization. The initial learning rate is set to 0.1, and we decay it by 0.1 and 0.01 at 20, 50 epochs. The weight decay is set to 5e-4. The whole training process contains 120 epochs. The $\lambda_{ml}$ and $\lambda_{ib}$ are set to 0.01 and 0.0005, respectively. Note that, since the partial re-id datasets only have single modality pedestrian images, we recycle the single modality to adapt the architecture in CMInfoNet. Our model is implemented with PyTorch.

\subsection{Ablation Study}
In this section, we evaluate the effectiveness of each component of our proposed method.

\textbf{Evaluation on Mutual Information Bottleneck.} We first evaluate how much improvement can be made by the mutual information bottleneck on SYSU-MM01 dataset under the all-search mode. It is worth noting that the mutual information bottleneck can be regarded as a Plug-and-Play block to existing methods. As shown in Table~\ref{table1}, we implement the AGW~\cite{agw} as a baseline and embed our mutual information bottleneck network into the feature extraction module to extract informative information. From the first two lines of Table~\ref{table1}, MIB brings $\textbf{4.24\%}$ Rank-1 and $\textbf{3.87\%}$ mAP increases compared with row 1 `Baseline'. Similar improvements can be observed in row 3 and 4, which suggests that information bottleneck not only guides the baseline network to focus on more informative parts of each modality but also filter out the noisy information. Note that, CMInfoNet*, CMInfoNet$\star$ and `Baseline + MIB + CMC' are optimized without $\mathcal{L}_{ml}$. The difference between CMInfoNet* and CMInfoNet$\star$ is that the former are trained without saliency search module.

\textbf{Evaluation on Cross-modality Consensus Module.} We also perform comparative experiments on CMC module, whose results are shown in Table~\ref{table1} and Table~\ref{table2}. The CMC module is also flexible and adaptable. The baseline obtains $\textbf{6.33\%}$ Rank-1 and $\textbf{6.43\%}$ mAP enhancements compared with row 3 in Table~\ref{table1}. By the way, the performance can gain further improvement with both the two modules added. In CMInfoNet, we treat one modality as the main modality and another as the auxiliary modality. The main modality is utilized to generate the  global modality-aware descriptor $\bm{\mathrm{G}}^{\mathrm{V}}$, while the auxiliary modality derives the cross-modality local-aware information $\bm{\mathrm{L}}^{\mathrm{I}}$ by a convolution layer. Therefore, we evaluate different kinds of combination of the main and auxiliary modalities in CMC module. The results are reported in Table~\ref{table2}. We can conclude that the the global operation can help obtain stable improvements on both main and auxiliary modalities compared with the results of the first row in Table~\ref{table2}. The results also reveal that global information is beneficial to guide the information redundancy recognition. At the same time, the combination of main modality and auxiliary modality can behave better. Both the global information and local characteristics can be integrated to detect the task-irrelevant information redundancy from the main modality.

\textbf{Evaluation on Training Objectives and Fusion Methods.} 
Since each loss item performs an important role in CMInfoNet to help learn pedestrian representations. To verify the effectiveness of each loss item, we perform a set of experiments and report the results in Table~\ref{table3}. We start with only the identity loss, the other loss items are added gradually. The results show that each loss item can help improve the performance of CMInfoNet. We also evaluate different similarity metrics in $\mathcal{L}_{ml}$, the results are shown in Table~\ref{table4}. We adopt three kinds of similarity metrics, i.e., Cosine similarity, Jensen-Shannon divergence (JS) and Mean Squared Error (MSE). The $\mathcal{L}_{ml}$ equipped with cosine similarity outperforms the MSE based $\mathcal{L}_{ml}$ by $\textbf{1.21\%}$ Rank-1 and $\textbf{0.23\%}$ mAP on SYSU-MM01, by $\textbf{0.17\%}$ Rank-1 and $\textbf{0.28\%}$ mAP on RegDB. The results on both two datasets also demonstrate the robustness of $\mathcal{L}_{ml}$.

\begin{table}[tbp]
\small
\centering
\caption{Performance comparison on Occluded-DukeMTMC. The methods in the 1st group are designed for the holistic re-id problem. The methods in the 2nd group utilize the pose information. The methods in the 3rd group are designed for the Partial Re-ID problem. The 4th group uses the transformer model for partial re-id. The 5th group is our method.}
\begin{tabular}{l|cccc}
\hline
Methods&Rank-1&Rank-5&Rank-10&mAP\\
\hline
\hline
LOMO+XQWDA~\cite{metriclearning1}&8.1&17.0&22.0&5.0\\
DIM~\cite{devil}&21.5&36.1&42.8&14.4\\
Part\ Aligned~\cite{part}&28.8&44.6&51.0&20.2\\
Random Erasing~\cite{randomerasing}&40.5&59.6&66.8&30.0\\
HACNN~\cite{hacnn}&34.4&51.9&59.4&26.0\\
Adver Occluded~\cite{adver}&44.5&-&-&32.2\\
PCB~\cite{pcb}&42.6&57.1&62.9&33.7\\
\hline
Part\ Bilinear~\cite{partbilinear}&36.9&-&-&-\\
FD-GAN~\cite{fdgan}&40.8&-&-&-\\
HOReID~\cite{HOReID}& 55.1 &-&-& 43.8 \\
PGFA~\cite{Occluded-Duke}&51.4&-&-&37.3\\
FED~\cite{FED}&\textbf{68.1}&-&-& \underline{56.4} \\
\hline
DSR~\cite{dsr}&40.8&58.2&65.2&30.4\\
SFR~\cite{sfr}&42.3&\underline{60.3}&\underline{67.3}&32.0\\
PVPM~\cite{PVPM}& 47 &-&-& 37.7 \\
\hline
PAT~\cite{PAT}& 64.5 &-&-& 53.6 \\
ViT Baseline~\cite{TransReID}&60.5 &-&-& 53.1\\
TransReID~\cite{TransReID}& 64.2 &-&-& 55.7 \\
\hline
CMInfoNet&\underline{67.3}&\textbf{82.2}&\textbf{87.5}&\textbf{56.8}\\
\hline
\end{tabular}
\label{table7}
\end{table}

\begin{table}[tbp]
\small
\centering
\caption{Performance comparison on Partial-REID and Partial\_iLIDS.}
\begin{tabular}{l|c|c|c|c}
\hline
\multirow{2}*{Method}&\multicolumn{2}{c|}{Partial-REID}&\multicolumn{2}{c}{Partial\_iLIDS}\\
\cline{2-5}
& Rank-1 & Rank-3 & Rank-1 & Rank-3\\
\hline
\hline
MTRC~\cite{mtrc}&23.7&27.3&17.7&26.1\\
AMC+SWM~\cite{Partial-REID}&37.3&46.0&21.0&32.8\\
DSR~\cite{dsr}&50.7&70.0&58.8&67.2\\
SFR~\cite{sfr}&56.9&\underline{78.5}&\underline{63.9}&\underline{74.8}\\
PAT~\cite{PAT}& \textbf{88.0} & - & -& - \\
ViT Baseline~\cite{TransReID}& 73.3 & 74.0 & - & - \\
TransReID~\cite{TransReID}& 71.3 & 68.6 & -& - \\
\hline
CMInfoNet&\underline{77.5}&\textbf{83.7}&\textbf{68.1}&\textbf{75.5}\\
\hline
\end{tabular}
\label{table8}
\end{table}

\begin{table}[tbp]
\small
\centering
\caption{Performance comparison on Occluded-REID.}
\begin{tabular}{l|cc}
\hline
Methods&Rank-1&mAP\\
\hline
PCB~\cite{pcb}& 41.3 & 38.9 \\
DSR~\cite{dsr}& 72.8 & 62.8 \\
FRR~\cite{FRR}& 78.3 & 68.0 \\
PVPM~\cite{PVPM}& 70.4 & 61.2 \\
HOReID~\cite{HOReID}& 80.3 & 70.2 \\
PAT~\cite{PAT}& 81.6 & 72.1 \\
ViT Baseline~\cite{TransReID}& 81.2 & 76.7 \\
TransReID~\cite{TransReID}& 70.2 & 67.3 \\
FED~\cite{FED}& \underline{86.3} & \underline{79.3} \\
\hline
CMInfoNet&\textbf{88.5}&\textbf{82.9}\\
\hline
\end{tabular}
\label{table9}
\end{table}

\subsection{Comparison with State-of-the-art Methods}
In this section, we compare our method with state-of-the-arts and report the results on holistic and partial person Re-ID datasets, respectively.

\textbf{Comparisons on Holistic Datasets.} Table~\ref{table5} demonstrates the experimental results on SYSU-MM01 dataset. Note that $*$ denotes that we report the results of \cite{ssft} under single-query setting for fair comparisons with other methods. The bold results represent the best result, and the underlined results are denoted as the sub-optimal. According to the results, the following observations can be made: 1) Our method performs much better than those time- and space-expensive image generation methods, which reduces intermediate steps and avoids introducing additional noise. 2) Compared with the method using both global and local features~\cite{ddag} as well as the NAS based method \cite{CM-NAS}, CMInfoNet significantly outperforms it by a large margin. Notably, CMInfoNet shows superior performance among other methods, indicating that the MIB, CMC module as well as the saliency search module are beneficial for practical pedestrian retrieval. 

The experiments on RegDB dataset (Table~\ref{table6}) suggest that CMInfoNet is robust against different query settings. Our method has distinct advantages over state-of-the-arts on RegDB. Under $\textit{Visible-to-Infrared}$ mode, CMInfoNet exceeds state-of-the-art CM-NAS~\cite{CM-NAS} by $\textbf{1.59\%}$ in Rank-1 and $\textbf{2.36\%}$ in mAP. The results are also improved when switching to $\textit{Infrared-to-Visible}$ mode, which further verifies the generalization ability of CMInfoNet. It performs well on both visible-to-infrared and infrared-to-visible settings by MIB and CMC modules. Through cross modality learning, the modality-invariant identity features can be learned better with more discriminative ability.

\textbf{Comparisons on Occluded Datasets.} The results on Occluded-DukeMTMC, Partial-REID, Partial\_iLIDS and Occluded-REID are shown in Table~\ref{table7},~\ref{table8} and~\ref{table9}. In Table~\ref{table7}, our method is compared with four kinds of methods, i.e., the holistic re-id problem, the methods that utilize the pose information, the partial re-id problem, and the transformer-based methods. It can be found that the performance of transformer-based methods have a great improvement through the transformer encoder-decoder structure. Our method have also achieved superior performance when compared to these methods. Since the FED~\cite{FED} is trained with the help of the pose information, it has achieved a higher accuracy. However, our method is still competitive with FED. 

In Table~\ref{table8}, we demonstrate the results on Partial-REID and Partial\_iLIDS datasets. PAT~\cite{PAT} conducts the diverse part discovery through the transformer, which have more specific feature detectors and are important to improve the network performance on occluded pedestrian images. We have achieved the highest performance on Partial\_iLIDS, and exceed the state-of-the-art SFR~\cite{sfr} by $\textbf{4.2\%}$ in Rank-1 and $\textbf{0.7\%}$ in Rank-3. Table~\ref{table9} shows the results on Occluded-REID. Our method has also achieved the highest performance when compared to the FED~\cite{FED} and the transformer-based methods~\cite{PAT,TransReID}.

\begin{figure}[htbp]
\centering
\subfloat[Baseline.]
{\includegraphics[width=0.482\linewidth]{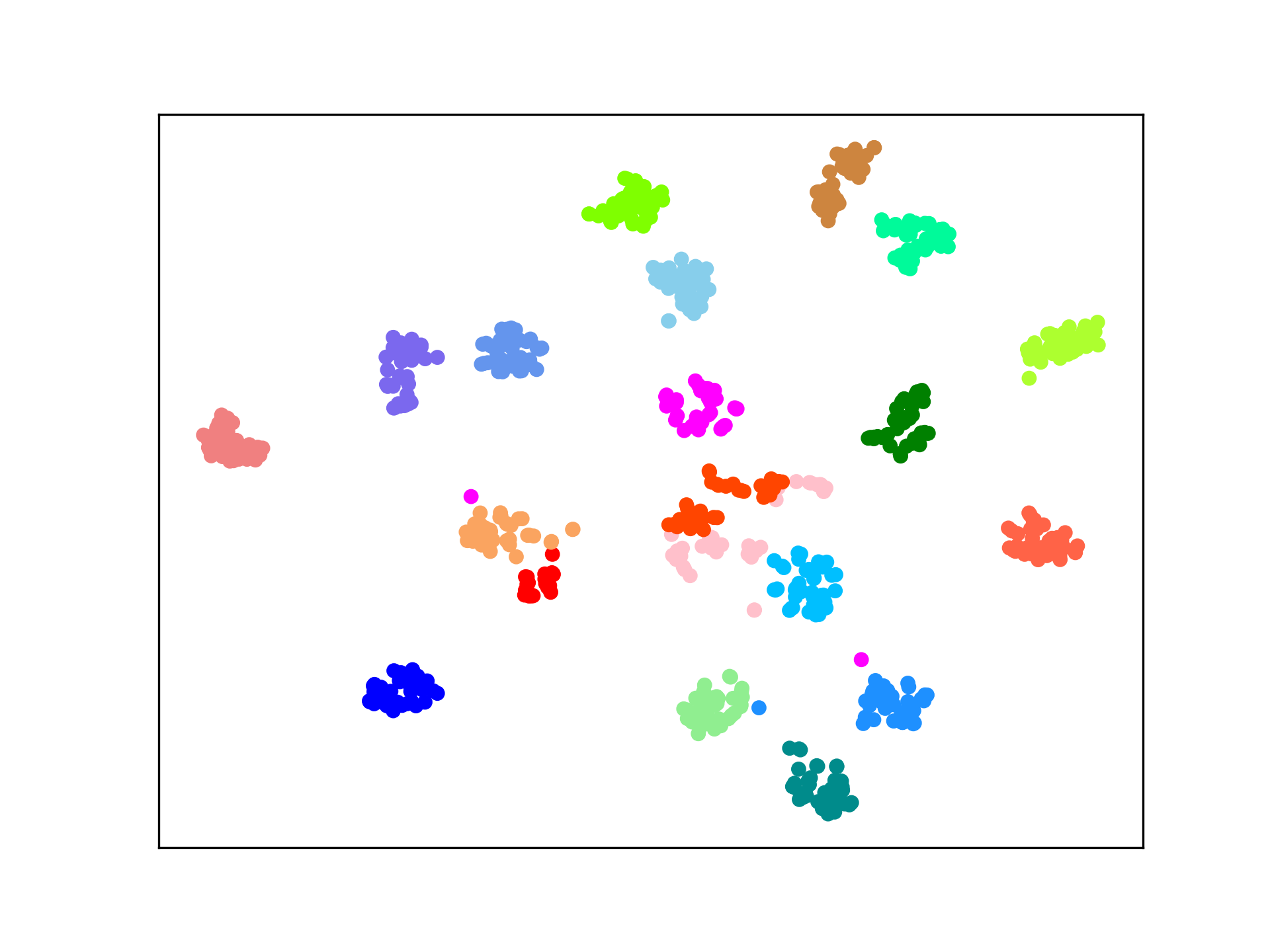}
}
\subfloat[CMInfoNet.]
{\includegraphics[width=0.482\linewidth]{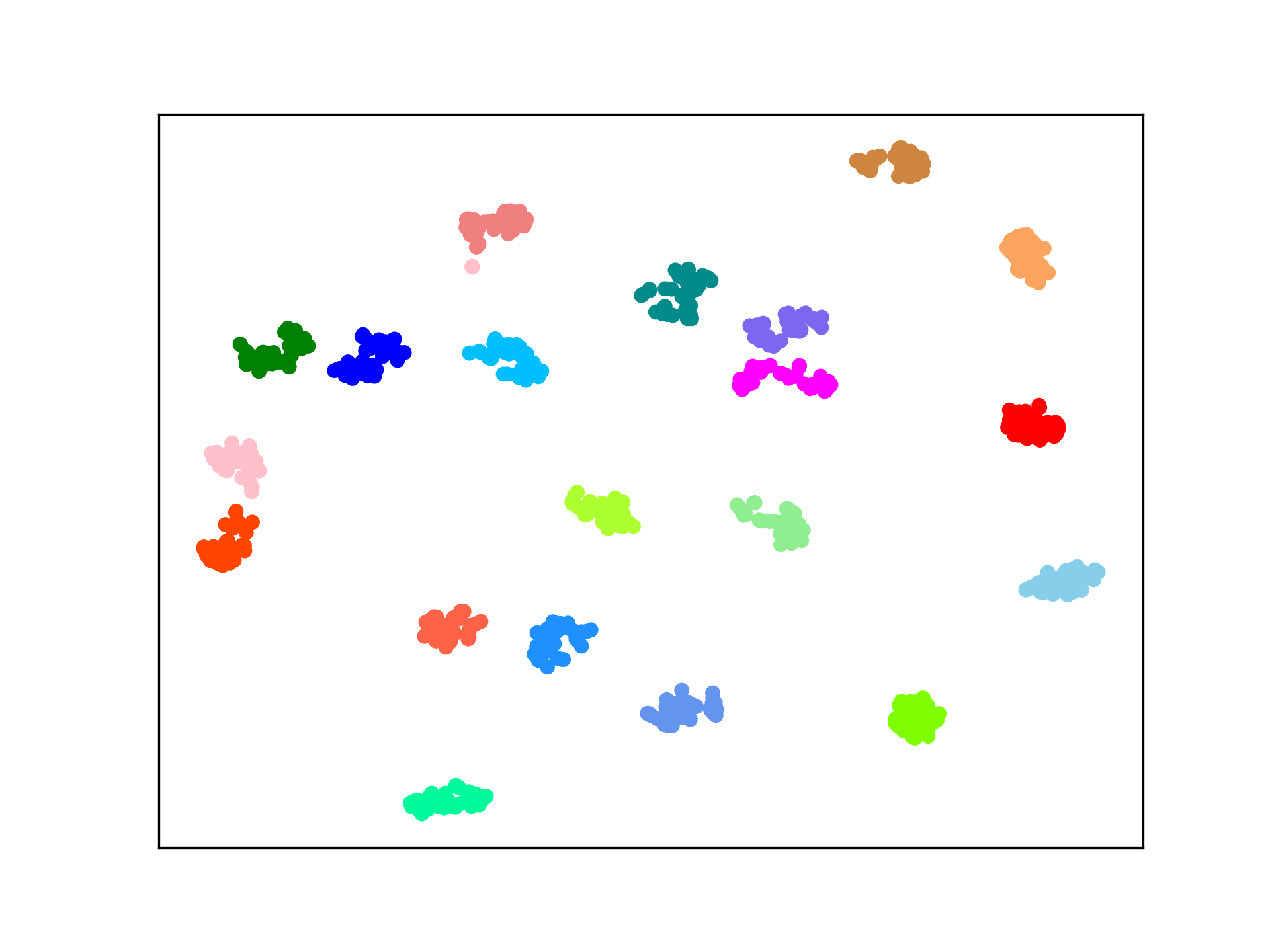}
}
\caption{The t-SNE visualization of features on SYSU-MM01 dataset. 20 identities of testing set are randomly selected. Different colors represent different identities.}
\label{tsne}
\end{figure}

\subsection{Visualization Analysis}
To further analyze the effectiveness of CMInfoNet, we use t-SNE~\cite{tsne} to transform high dimensional features vectors into two-dimensional vectors. As shown in Fig.~\ref{tsne}, compared to the visualization results of baseline, the features extracted from CMInfoNet are better clustered together. The distances between the centers and boundaries among different identities are more obvious, verifying that our method is more discriminative.

\section{Conclusion}
In this paper, we propose a novel mutual information based cross-modality consensus network, namely CMInfoNet. The modality-invariant identity features with the most representative information are supposed to be extracted by optimizing the mutual information bottleneck trade-off. For the model efficiency and precision, an saliency search module is proposed to find the most prominent parts of pedestrians. The information redundancies are reduced at the same time. Moreover, a cross-modality consensus module is designed to align the visible and infrared modalities and complement each other to explore cross- and intra-modality cues. Accompanied with a modality contrastive loss, the inter-modality consistency could be further explored. Extensive experiments have demonstrated the superior performance of the proposed CMInfoNet.

\bibliographystyle{IEEEtran}
\bibliography{ref}

\end{document}